\documentclass[lettersize,journal]{IEEEtran}
\usepackage{amsmath,amsfonts}
\usepackage{algorithmic}
\usepackage{algorithm}
\usepackage{array}
\usepackage[caption=false,font=normalsize,labelfont=sf,textfont=sf]{subfig}
\usepackage{textcomp}
\usepackage{stfloats}
\usepackage{url}
\usepackage{verbatim}
\usepackage{graphicx}
\usepackage{cite}
\usepackage{csquotes}
\usepackage{booktabs}
\usepackage{xcolor}
\usepackage[table,xcdraw]{xcolor}
\usepackage{multirow}
\begin{document}

\title{A Physically-Grounded Attack and Adaptive Defense Framework for Real-World Low-Light Image Enhancement}

\author{Tongshun Zhang, Pingping Liu, Yuqing Lei, Zixuan Zhong, Qiuzhan Zhou and Zhiyuan Zha, \emph{Senior Member, IEEE}
        % <-this % stops a space
\thanks{
%This work was supported by Jilin Province Industrial Key Core Technology Tackling Project (20230201085GX).

Tongshun Zhang, Pingping Liu, and Yuqing Lei are in the College of Computer Science and Technology, Jilin University, Changchun 130015, P.R. China and the Key Laboratory of Symbolic Computation and Knowledge Engineering of Ministry of Education, Jilin University, Changchun 130015, P.R. China (email: tszhang23@mails.jlu.edu.cn; liupp@jlu.edu.cn; leiyq25@ mails.jlu.edu.cn).  \emph{(Corresponding author: Pingping Liu.)}

Zixuan Zhong is in the College of Software, Jilin University, Changchun 130015, P.R. China (email: zhongzx24@mails.jlu.edu.cn).

Qiuzhan Zhou and Zhiyuan Zha are in the College of Communication Engineering, Jilin University, Changchun 130015, P.R. China (email: zhouqz@jlu.edu.cn; zhiyuan\_zha@jlu.edu.cn).
}% <-this % stops a space
}

\markboth{Journal of \LaTeX\ Class Files,~Vol.~14, No.~8, August~2021}%
{Shell \MakeLowercase{\textit{et al.}}: A Sample Article Using IEEEtran.cls for IEEE Journals}


\maketitle

\begin{abstract}
Limited illumination often causes severe physical noise and detail degradation in images. Existing Low-Light Image Enhancement (LLIE)  methods frequently treat the enhancement process as a blind black-box mapping, overlooking the physical noise transformation during imaging, leading to suboptimal performance. To address this, we propose a novel LLIE approach, conceptually formulated as a physics-based attack and display-adaptive defense paradigm. Specifically, on the attack side, we establish a physics-based Degradation Synthesis (PDS) pipeline. Unlike standard data augmentation, PDS explicitly models Image Signal Processor (ISP) inversion to the RAW domain, injects physically plausible photon and read noise, and re-projects the data to the sRGB domain. This generates high-fidelity training pairs with explicitly parameterized degradation vectors, effectively simulating realistic attacks on clean signals. On the defense side, we construct a dual-layer fortified system. A noise predictor estimates degradation parameters from the input sRGB image. These estimates guide a degradation-aware Mixture of Experts (DA-MoE), which dynamically routes features to experts specialized in handling specific noise intensities. Furthermore, we introduce an Adaptive Metric Defense (AMD) mechanism, dynamically calibrating the feature embedding space based on noise severity, ensuring robust representation learning under severe degradation. Extensive experiments demonstrate that our approach offers significant plug-and-play performance enhancement for existing benchmark LLIE methods, effectively suppressing real-world noise while preserving structural fidelity. The sourced code is available at \url{https://github.com/bywlzts/Attack-defense-llie}.
\end{abstract}


\begin{figure}[!t]
  \centering
  \includegraphics[width=1.0\linewidth]{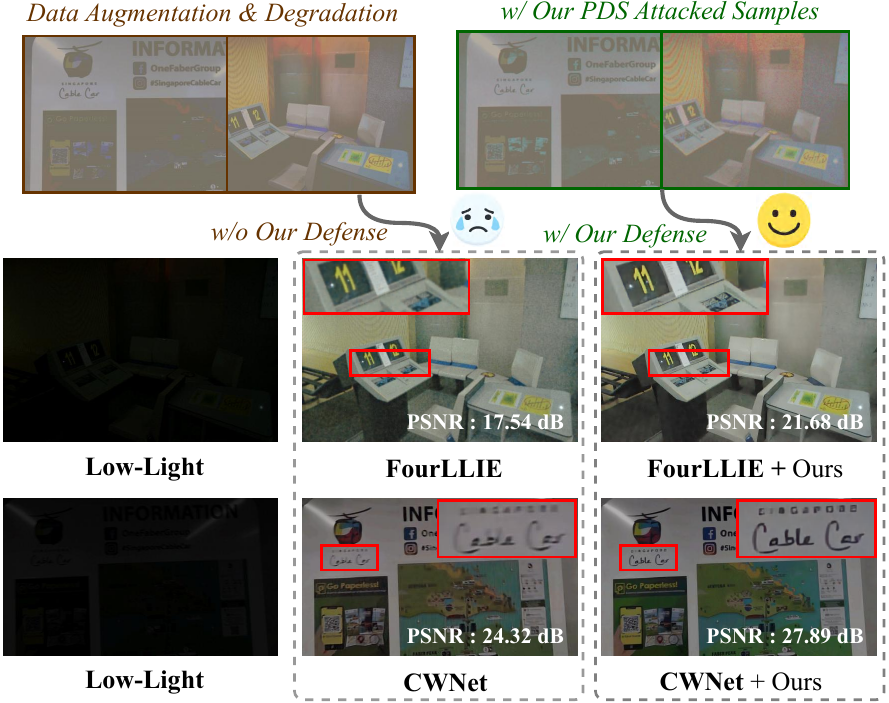} 
  \vspace{-0.6cm}
  \caption{
  Visual comparison of existing LLIE methods against our proposed approach. While current methods demonstrate good brightness recovery, they often suffer from subpar noise suppression, as highlighted by the noisy regions. Our proposed adversarial framework demonstrably provides superior noise reduction while maintaining structural fidelity.
  }
  \vspace{-0.6cm}
  \label{fig:page_one_vis}
\end{figure}

\begin{figure*}[!t]
  \centering
  \includegraphics[width=1.0\linewidth]{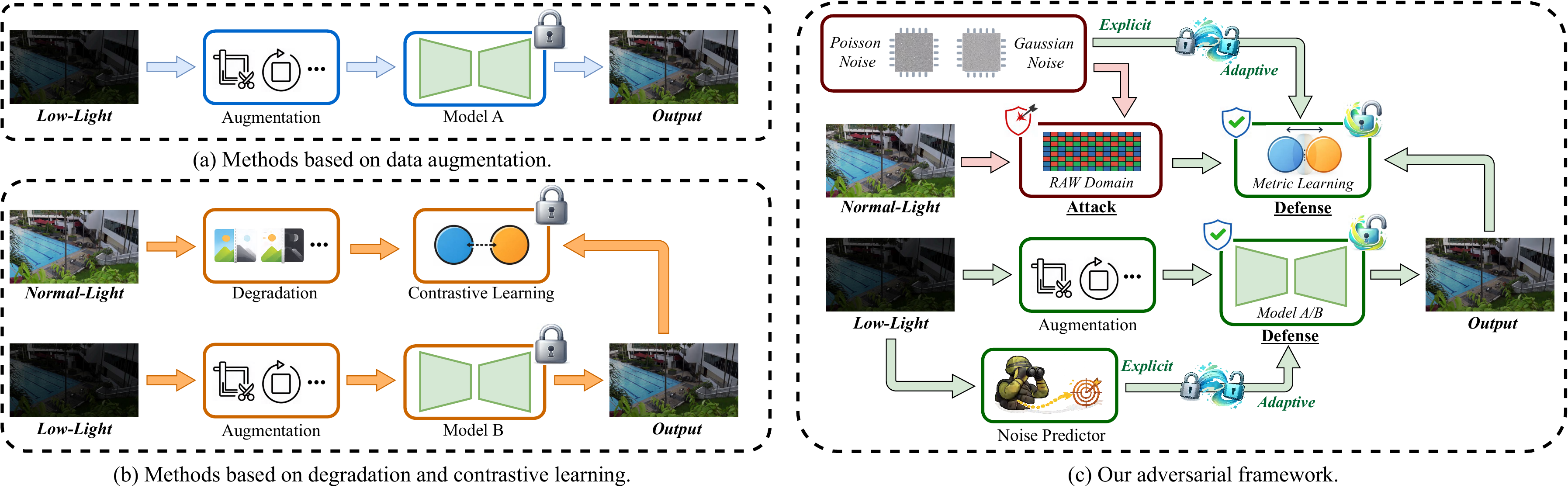} 
  \vspace{-0.6cm}
  \caption{
Comparison of different LLIE training paradigms. \textbf{(a) Data augmentation-based methods} treat LLIE as a passive black-box mapping from augmented low-light inputs. \textbf{(b) Degradation + contrastive learning methods} generate negative samples via handcrafted degradations, but often overlook the physical nature of sensor noise and lack adaptation to degradation severity. \textbf{(c) Our adversarial framework} employs a physics-guided RAW domain attack for plausible noise and explicit priors. A sentinel noise predictor offers explicit noise awareness, while adaptive architectural and metric defenses respond to attack severity for targeted noise suppression and robust enhancement.
  }
  \vspace{-0.5cm}
  \label{fig:compare_current_method_vis}
\end{figure*}

\section{Introduction}

\IEEEPARstart{C}{apturing} high-fidelity images in low-light environments has been a persistent challenge in the field of computational photography. Due to the limitation of photon scarcity, the signals captured by sensors are inherently weak, leading to severely constrained signal-to-noise ratios (SNR). As a result, the acquired images suffer from dual degradation: they are not only blurry due to low visibility but also affected by complex, signal-dependent noise. This degradation extends beyond perceptual fidelity; the loss of high-frequency details, such as fine textures and edges, adversely impacts downstream high-level visual tasks.

To address this ill-posed problem, early methods primarily relied on statistical priors and signal processing theories. Techniques based on Retinex theory~\cite{guo2016lime}, gamma correction~\cite{huang2012efficient}, and histogram equalization~\cite{rahman2016adaptive} focused mainly on dynamic range adjustment. However, these methods often inadvertently amplify potential artifacts during contrast stretching, as they overlook noise distributions. On the other hand, while traditional model-based denoising algorithms~\cite{elad2006image, mairal2009non, dabov2008image} established rigorous mathematical frameworks, they typically assume idealized noise models, which makes them ineffective at addressing the inherent spatially varying noise characteristics present in low-light scenes.

The advent of deep learning has significantly advanced LLIE methods~\cite{kong2023efficient, mao2023intriguing, yan2023sharpformer, retinexformer}, achieving remarkable brightness recovery. However, a persistent challenge remains: many state-of-the-art LLIE approaches, while effective at brightening images, still exhibit subpar noise suppression, especially in real-world scenarios. As vividly illustrated in Fig.~\ref{fig:page_one_vis}, methods like FourLLIE~\cite{wang2023fourllie} and CWNet~\cite{zhang2025cwnet} demonstrate this limitation, often failing to effectively handle complex, signal-dependent noise. This deficiency largely stems from current LLIE training paradigms' reliance on generic data augmentation in the sRGB space. These augmentations implicitly treat degradations as simple pixel-wise perturbations, critically overlooking the intricate physical camera image formation pipeline and its influence on noise characteristics. Consequently, the synthesized noise frequently fails to capture the signal-dependent, non-linear, and spatially/cross-channel correlated nature of real low-light noise. This makes disentangling and suppressing entangled noise post-illumination boosting particularly difficult. A common practice to mitigate this is sequential processing, where enhancement and denoising are performed separately. Yet, this divide-and-conquer strategy is prone to error propagation: enhancement can amplify noise, while pre-denoising might over-smooth subtle details in dark regions, hindering their subsequent recovery. This critical trade-off highlights the urgent need for a unified LLIE framework that jointly optimizes illumination recovery and noise suppression, grounded in a deeper understanding of physical degradation processes. Such a framework could revolutionize how LLIE methods are trained and deployed, moving beyond heuristic data augmentation towards physics-informed simulation.

Despite the architectural advancements achieved by the aforementioned methods, a fundamental bottleneck remains.
Fig.~\ref{fig:compare_current_method_vis}(a) summarizes the dominant augmentation-driven paradigm for LLIE~\cite{wang2023fourllie, zamir2022restormer, wang2022uformer, zhou2022lednet}: models are trained in the sRGB space, primarily utilizing label-preserving data augmentations (e.g., random cropping, flipping, rotation, translation, and scaling) to enrich the training distribution and improve generalization, while learning a direct mapping from low-light inputs to normal-light targets. In contrast, Fig.~\ref{fig:compare_current_method_vis}(b) depicts a degradation- and contrastive-learning-based paradigm~\cite{zhang2025cwnet, yang2023implicit, guo2024onerestore}. This approach introduces an additional degradation branch applied to normal-light images (e.g., blur, tone shift, brightness adjustment, and synthetic noise) to explicitly construct negative (or hard) samples; a contrastive objective then encourages the enhanced outputs to be dissimilar from these degraded instances in representation space. Notably, augmentation and degradation differ in essence: the former seeks task-invariant transformations that preserve desired target semantics and mainly increases sample diversity, whereas the latter applies task-variant corruptions that intentionally reduce image quality and are used to organize positives/negatives for contrastive learning. Nevertheless, both paradigms typically operate after the ISP pipeline in the sRGB domain and thus still treat LLIE largely as a passive black-box mapping. This overlooks that the ISP pipeline (white balance, demosaicing, color correction, and tone mapping) applies a cascade of nonlinear operations that fundamentally reshapes sensor noise, yielding artifacts that are non-Gaussian, signal-dependent, and spatially/chromatically correlated in sRGB. Moreover, degradation-based contrastive schemes often treat all degraded views as uniformly negative, without modeling the continuous spectrum of degradation severity, which weakens the ability to distinguish hard negatives from easy ones and ultimately limits enhancement performance. These observations raise a critical question: \textbf{\emph{how can existing LLIE methods overcome their inherent limitations in noise suppression and break free from the implicit entanglement of noise in the sRGB domain?}}

To circumvent the difficulty of modeling ISP-induced artifacts, recent studies \cite{jiang2025learning, yang2025learning, li2025noise} have advocated shifting the LLIE paradigm to the raw sensor (RAW) domain. In this domain, noise follows physically interpretable distributions—typically exhibiting a linear dependency on signal intensity—making it more amenable to mathematical modeling. Pioneering works \cite{wang2020practical, wei2021physics, zhang2021rethinking, lu2022progressive} have demonstrated that accurate noise calibration in the RAW domain leads to substantially improved denoising fidelity. Building upon this insight, \cite{kousha2022modeling} further employed normalizing flows to explicitly model the complex posterior distribution of camera noise. Despite the clear advantage of accurate noise modeling in the RAW domain, a substantial gap remains between these approaches and real-world deployment. The vast majority of consumer devices and practical vision systems still operate exclusively on sRGB images (e.g., JPEG or HEIC formats), primarily due to the prohibitive storage and transmission costs associated with uncompressed RAW data \cite{zhang2021rethinking, li2025noise}. This reality motivates a second fundamental question: \textbf{\emph{how can we exploit the physical tractability of RAW-domain noise modeling while adhering to the practical constraint that real-world applications predominantly rely on sRGB inputs?}}

To address these aforementioned challenges and bridge the gap between physically accurate RAW-domain modeling and practical sRGB deployment, as shown in Fig.~\ref{fig:compare_current_method_vis}(c), we explicitly revisit the physical origins of low-light degradation and introduce a novel adversarial paradigm: the physics-based attack and display-adaptive defense framework. Our core motivation is that while many LLIE methods achieve global brightness enhancement, their performance severely degrades under entangled and signal-dependent noise. This degradation is exacerbated in the sRGB domain, where the cascade of nonlinear ISP operations fundamentally reshapes sensor noise, rendering conventional data augmentation or handcrafted sRGB degradations insufficient for reliable noise suppression. Our proposed framework is built upon three complementary perspectives to systematically tackle these issues.

\textbf{\emph{Why an adversarial paradigm?}}
An adversarial perspective helps identify and control the root causes of failure in existing LLIE methods. Instead of passively learning a black-box mapping from noisy sRGB inputs to clean outputs, we cast enhancement as a game between a goal-driven physical attacker and a noise-aware defender. Crucially, the ``adversary'' here is not a gradient-based attacker; rather, it generates physically rigorous ``worst-case scenarios'' that remain plausible under real imaging physics, thereby encouraging robustness against real-world low-light degradations.

\textbf{\emph{What to attack and how to attack?}}
Given that current methods already achieve satisfactory illumination recovery, the key vulnerability lies in noise modeling. Due to the nonlinear ISP pipeline, noise in the sRGB domain becomes highly entangled and difficult to characterize. Therefore, we attack the noise distribution in the linear RAW domain, where noise follows explicit physical laws and can be parameterized analytically. On the attack side, we actively synthesize ``hard'' training samples using a physics-based Degradation Synthesis (PDS) module. Unlike black-box adversarial training (e.g., GAN-based artifact hallucination), PDS analytically inverts the ISP pipeline to RAW, injects physically plausible photon (shot) noise and read noise governed by sensor statistics, and then re-projects the corrupted signal back to sRGB through a forward ISP. This procedure fundamentally differs from conventional augmentation: PDS does not produce heuristic perturbations, but instead generates explicit degradation priors in the form of noise-parameter vectors that quantify the type and severity of realistic noise. As a result, enhancement is transformed from blind estimation into an informed attack--defense process in which noise-related vulnerabilities are explicitly exposed.

\textbf{\emph{How to construct the defense?}}
On the defense side, we equip the enhancement network with explicit noise awareness. Architecturally, we introduce a Noise Predictor that estimates degradation parameters directly from the attacked sRGB input, providing actionable ``intelligence'' for subsequent enhancement. These estimates then guide a degradation-aware Mixture of Experts (DA-MoE). Leveraging the Central Limit Theorem (CLT), DA-MoE performs dynamic feature routing: for lower-resolution features where aggregated noise tends to converge to Gaussian-like distributions, the gating prioritizes Gaussian-oriented experts; for high-resolution features where signal-dependent shot noise remains prominent, the gating emphasizes Poisson-oriented experts and modulates them via Spatial Feature Transform (SFT) to better accommodate intensity-dependent corruption. Finally, to avoid the optimization settling into suboptimal local minima on heavily attacked samples, we propose an Adaptive Metric Defense (AMD) that calibrates feature-space margins according to attack severity, enabling stronger suppression for hard negatives while preserving fine-grained structures for mild degradations.

By coupling physics-based attacks with an adaptive and explicit defense system, our proposed physics-based attack and display-adaptive defense framework establishes a principled interaction between degradation modeling and feature enhancement. This adversarial process alleviates the implicit noise entanglement induced by the sRGB ISP pipeline, enabling the network to learn noise-discriminative and illumination-robust representations, thereby leading to superior enhancement performance under real-world low-light conditions.
In summary, our main contributions are as follows:
\begin{itemize}
    \item We identify the fundamental limitation of existing LLIE methods as the implicit entanglement of noise in the sRGB domain, and reformulate LLIE as an adversarial problem between physics-based attacks and noise-aware defenses.
    
    \item We propose a physics-based Degradation Synthesis (PDS) pipeline that explicitly targets sensor noise. By analytically inverting and re-applying the ISP pipeline, PDS synthesizes physically plausible degradations, exposes noise-related failure modes, and provides explicit, interpretable noise priors.
    
    \item We introduce a dual-layer explicit and adaptive defense system, comprising a degradation-aware Mixture of Experts (DA-MoE) and an Adaptive Metric Defense (AMD) mechanism. This system exploits attack-aware noise information at both architectural and metric levels, enabling targeted noise purification and dynamically calibrated optimization.
    
    \item Extensive experiments on real-world low-light benchmarks demonstrate that our proposed adversarial framework consistently improves robustness and generalization, offering a new perspective for noise-aware LLIE.
\end{itemize}

\begin{figure*}[!t]
  \centering
  \includegraphics[width=1.0\linewidth]{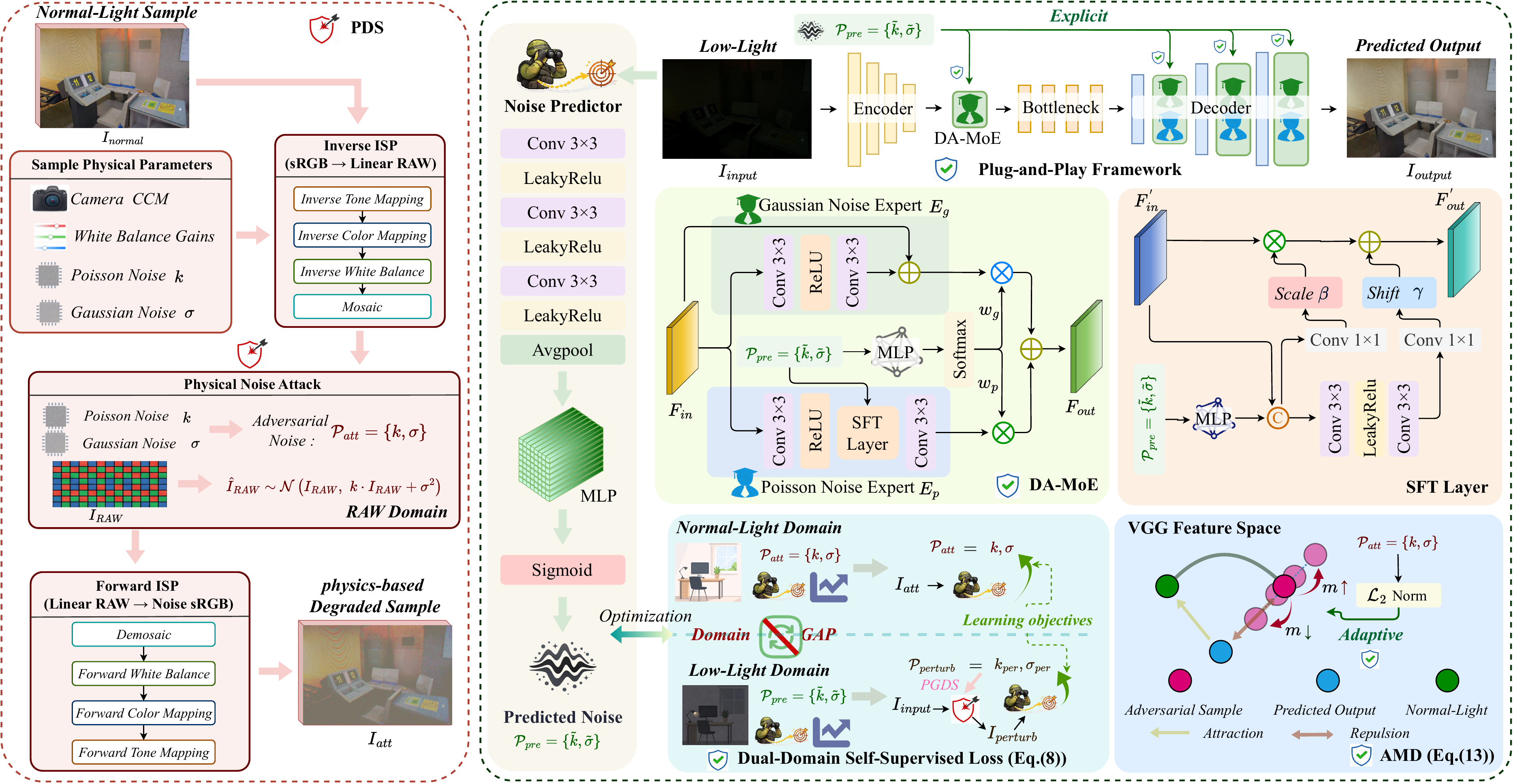} 
  \vspace{-0.5cm}
  \caption{
    Overview of the proposed \textbf{physics-based adversarial framework}.
    \textbf{Attack stream (left, red):} \textbf{PDS} analytically inverts clean sRGB images to the linear RAW domain, injects Poisson and Gaussian noise parameters $(k,\sigma)$ for sensor-level attacks, and re-projects the corrupted signal to sRGB via a forward ISP to generate realistic physics-based degraded samples with ground-truth physical priors ($\mathcal{P}_{att}$).
    \textbf{Defense stream (right, green):} A sentinel Noise Predictor estimates degradation parameters ($\mathcal{P}_{pre}$) from the low-light input and uses them to gate/modulate \textbf{DA-MoE} blocks and SFT layers, trained with a \textbf{Dual-Domain Self-Supervised Loss}. The \textbf{DA-MoE} is plug-and-play for existing \textbf{enhancement} backbones, and the AMD further adjusts the margin $m$ according to the injected noise parameters ($\mathcal{P}_{att}$) to maintain appropriate feature distances under varying degradation severities.
  }
  \vspace{-0.3cm}
  \label{fig:overall}
\end{figure*}

\section{Related Work}

\subsection{Low-Light Image Enhancement (LLIE)}
LLIE aims to improve image quality under suboptimal lighting conditions by enhancing brightness/contrast, suppressing noise, and recovering fine details. Benefiting from data-driven deep learning, recent LLIE methods can be broadly categorized and reviewed from the following perspectives:

\subsubsection{Spatial Domain Architectures}
Early deep LLIE methods primarily relied on Convolutional Neural Networks (CNNs). Approaches like Zero-DCE~\cite{guo2020zero} cast enhancement as curve estimation for dynamic range adjustment. Retinex-inspired designs, such as PairLLIE~\cite{fu2023learning} and Retinexformer~\cite{cai2023retinexformer}, introduce decomposition priors to learn adaptive reflectance or guide non-local interactions. To alleviate color shifts, CIDNet~\cite{yan2025hvi} utilizes an HVI color space with a U-Net-like architecture for luminance/chromaticity mappings. For larger receptive fields, IAGC~\cite{wang2023low} proposes hierarchical Transformer attention to infer dark-region pixels from long-range context. Furthermore, methods exploring external or implicit priors, including GLARE~\cite{zhou2024glare}, PercepLIE~\cite{wang2024perceplie}, and URWKV~\cite{xu2025urwkv}, employ codebook-based strategies or unified receptive-field mechanisms for global correction and local detail recovery. Auxiliary-modality priors have also guided enhancement, with SKF~\cite{wu2023learning} fusing semantic priors and SMG-LLE~\cite{xu2023low} generating structure/edge priors to steer brightness enhancement.

\subsubsection{Frequency Domain Approaches}
Frequency-based methods decompose images into complementary components, enabling targeted illumination enhancement while controlling noise and maintaining efficiency. \textbf{Fourier-based} techniques exploit distinct roles of amplitude and phase, with FourLLIE~\cite{wang2023fourllie} characterizing amplitude as low-frequency and phase as high-frequency for global information extraction. UHDFour~\cite{li2023embedding} leverages cross-resolution amplitude similarity for ultra-high-definition LLIE, and DMFourLLIE~\cite{zhang2024dmfourllie} incorporates infrared modality, highlighting phase processing in multi-modal fusion. \textbf{Wavelet-based} approaches, like Wave-Mamba~\cite{zou2024wave} and SPJFNet~\cite{zhang2025spjfnet}, improve efficiency through lossless downsampling and sub-band modeling to separate noise and preserve textures. DiffLL~\cite{jiang2023low} applies diffusion to restore wavelet low-frequency components.

\subsubsection{Generative and Vision-Language Models}
The advent of diffusion models and vision-language/foundation models has further advanced LLIE by improving generative fidelity and leveraging strong priors~\cite{yang2023implicit, morawski2024unsupervised, wu2024jores}. BiDiff~\cite{he2025degradation} introduces bidirectional diffusion optimization for jointly modeling low-/normal-light degradation, and LightenDiffusion~\cite{jiang2024lightendiffusion} combines Retinex decomposition with diffusion for unsupervised enhancement. Moving toward foundation-model synergy, FoCo~\cite{gu2025improving} proposes a self-supervised framework to improve perceptual quality and downstream performance. GPP-LLIE~\cite{zhou2025low} further leverages Vision-Language Model (VLM)-generated generative perception priors, providing global and local guidance through multiple visual attributes.

\subsubsection{RAW Domain LLIE}
Beyond sRGB-based pipelines, RAW-domain research has gained increasing attention due to its physically interpretable noise characteristics. Brooks et al.~\cite{brooks2019unprocessing} proposed an efficient noise synthesis procedure driven by sensor noise models to improve low-light denoising. SIED~\cite{jiang2025learning} builds a paired synthesis pipeline to generate RAW images under different illuminations for extremely dark scenes. ParamISP~\cite{kim2023paramisp} learns forward/inverse transforms between RAW and sRGB, enabling blur synthesis and RAW-domain LLIE. Nevertheless, these approaches often remain largely black-box in restoration, with limited explicit interaction between physically grounded noise modeling and the enhancement network.

\subsubsection{Joint LLIE and Deblurring}
Joint LLIE and deblurring addresses both low-light conditions and motion blur. LEDNet~\cite{zhou2022lednet} pioneered this line by introducing the LOL-Blur dataset and an end-to-end network for joint brightening and deblurring. ELEDNet~\cite{kim2024towards} further collects real-world data guided by event cameras, exploiting temporal cues to recover brightness and structure. To incorporate strong priors, VQCNIR~\cite{zou2024vqcnir} matches degraded inputs to high-quality codebook features, while FDN~\cite{tu2025fourier} analyzes brightness/noise via Fourier amplitude–phase decomposition and proposes Fourier decoupling self-attention.

\subsection{Attack and Defense Mechanisms}
Attack and Defense paradigms, originating from adversarial robustness, are broadly adopted in computer vision to enhance model resilience against various perturbations. SCOUT~\cite{yan2025scout} employs adversarial augmentation for improved data efficiency in camouflaged object detection, while OS-RFS~\cite{wang2025robust} leverages adversarial positive attacks for robust multi-modal learning. Similarly, A$^2$RNet~\cite{li2025a2rnet} integrates adversarial training with anti-attack losses to maintain image quality robustly.

However, despite their success in other vision tasks, this paradigm remains critically underexplored in image restoration and particularly in LLIE, especially concerning the physical nature of degradation. While existing attempts~\cite{guo2024onerestore, zhang2025cwnet} create negative samples via degradations or interference for augmentation or loss design, they fundamentally lack a unified framework where the physical attributes of the ``attack'' explicitly and systematically guide the ``defense'' mechanism, especially in understanding and mitigating complex, signal-dependent noise. This crucial gap, specifically how to leverage physically plausible degradation to inform a targeted defense for robust LLIE, is what our work aims to bridge.

\section{Method}

\subsection{Overview}
\label{subsec:overview}

As illustrated in Fig.~\ref{fig:overall}, we propose a novel physics-based adversarial defense framework for joint LLIE. Unlike conventional paradigms that formulate image enhancement as a passive black-box regression problem, our approach explicitly coordinates the interaction between a physics-based attack mechanism and an adaptive, explicit defense system. The proposed framework consists of two parallel yet tightly coupled streams.

\noindent \textbf{Physical Attack Stream.} 
To bridge the gap between synthetic and real-world degradations, we introduce a PDS module. Instead of relying on black-box neural rendering, PDS analytically inverts the ISP pipeline, mapping a clean normal-light image $I_{normal}$ from the sRGB domain back to the linear RAW domain via an inverse ISP process. In the RAW domain, we explicitly inject mixed Poisson and Gaussian noise, forming a physical attack vector $\mathcal{P}_{att} = \{k, \sigma\}$. The attacked RAW data are then re-projected to the sRGB domain through a forward ISP, producing physics-based degraded samples $I_{att}$.

\noindent \textbf{Explicit Adaptive Defense Stream.} 
The defense stream comprises three key components. First, a lightweight Noise Predictor acts as a sentinel, estimating noise parameters $\mathcal{P}_{pre}$ from the low-light input $I_{input}$. These predicted parameters are explicitly used as gating and modulation signals for the DA-MoE modules and SFT layers. Since the physical noise attacks are generated in the normal-light domain and may not be directly aligned with noise characteristics in the low-light domain, we further propose a dual-domain self-supervised loss to optimize the Noise Predictor. 
Second, the structural defense is instantiated by the DA-MoE, which is integrated into existing enhancement frameworks in a plug-and-play manner, being inserted before the bottleneck layer and after each upsampling layer in the decoder to produce the restored output $I_{output}$. The DA-MoE consists of a residual Gaussian noise expert and a Poisson noise expert modulated by SFT layers, enabling targeted denoising under different noise conditions. 
Finally, a metric-level defense, the AMD, is introduced at the optimization stage. This AMD mechanism adaptively calibrates the discriminative margin $m$ via the $\ell_2$ norm of the adversarial noise parameters $\mathcal{P}_{att}$, ensuring a precise and dynamic safety distance between the enhanced features and physics-based degraded samples.

\subsection{\textbf{Physics-based Degradation Synthesis (PDS)}}
\label{subsec:attack}

To bridge the gap between the theoretical optimality of RAW-domain processing and the practical prevalence of sRGB images, one might consider employing deep neural networks to invert the ISP pipeline. However, the mapping from sRGB to RAW is an ill-posed one-to-many problem, causing learning-based approaches~\cite{zamir2020cycleisp, kim2023paramisp} to frequently misinterpret high-frequency details or overfit to specific ISP implementations. Methods based on generative adversarial networks (GANs)~\cite{goodfellow2014generative} attempt to alleviate the domain gap, but typically operate as black boxes, suffering from mode collapse and producing hallucinated artifacts that violate sensor physics~\cite{guo2019toward}. Moreover, the synthesized noise lacks explicit parameterization, making it unsuitable for guiding downstream adaptive modules.

To overcome these limitations, we introduce the PDS module, as shown in Fig.~\ref{fig:overall} (left). The Poisson--Gaussian (PG) distribution is a canonical noise model that characterizes shot noise and read noise using Poisson and Gaussian distributions, respectively. With advances in sensor design and manufacturing, other signal-dependent noise sources are generally negligible. Accordingly, we simulate an inverse ISP pipeline to map a clean normal-light image $I_{normal}$ from the sRGB domain back to the linear RAW domain. In the RAW domain, we explicitly inject mixed Poisson and Gaussian noise, forming a physical attack vector $\mathcal{P}_{att} = \{k, \sigma\}$. The attacked RAW data are then re-projected to the sRGB domain via a forward ISP, producing physics-based degraded samples $I_{att}$. This process consists of three physically principled stages:

\noindent \textbf{Stage I: Physical Parameter Sampling and ISP Inversion.}
Given a clean normal-light sRGB image $I_{normal}$, we first sample a set of physically interpretable camera parameters to simulate diverse real-world imaging conditions. Specifically, we randomly select a camera color correction matrix (CCM) from a pool of real camera metadata~\cite{brooks2019unprocessing} (Canon 5D, Nikon D700, and Sony Alpha series), and sample channel-wise white balance gains $\mathbf{g}_{wb}$. Meanwhile, noise-related parameters are sampled to define the subsequent attack strength, including the shot noise coefficient $k$ and read noise standard deviation $\sigma$, which will be used in the RAW-domain physical attack.

With the sampled camera parameters, we analytically invert the ISP pipeline to recover a linear sensor-space representation. The inverse process consists of three steps:  
(i) inverse tone mapping using a gamma expansion $\Gamma^{-1}(\cdot)$ with $\gamma = 2.2$, which approximates the inverse of standard sRGB compression;  
(ii) inverse color correction via the inverse CCM $\mathbf{M}_{ccm}^{-1}$ to map linear sRGB back to the camera RGB space; and  
(iii) inverse white balance normalization by dividing each channel by its corresponding gain in $\mathbf{g}_{wb}$.  
Finally, a mosaicking operator $\mathcal{M}_{bayer}(\cdot)$ rearranges the linear camera RGB signal into an RGGB Bayer pattern, yielding a clean RAW measurement:
\begin{equation}
I_{RAW} = \mathcal{M}_{bayer}
\left(
\frac{\mathbf{M}_{ccm}^{-1} \cdot \Gamma^{-1}(I_{normal})}{\mathbf{g}_{wb}}
\right).
\end{equation}
This ISP inversion process relies solely on physically meaningful operations, avoiding the ambiguity and instability introduced by learned ISP inversion networks.

\begin{figure}[!t]
  \centering
  \includegraphics[width=1.0\linewidth]{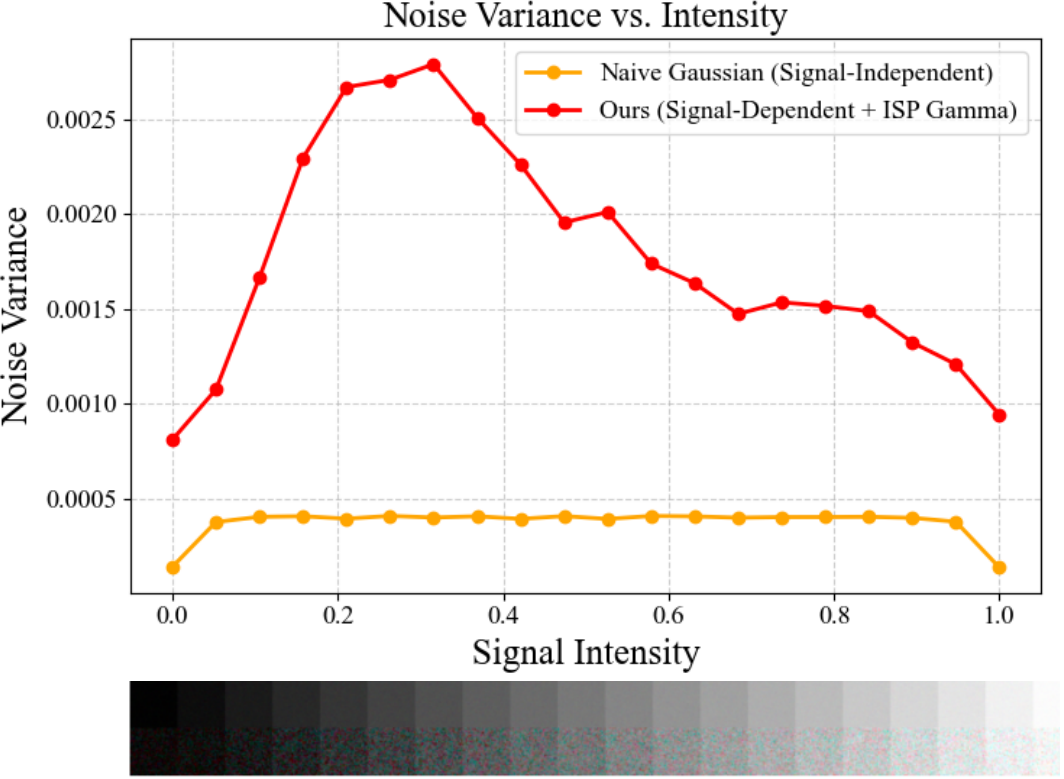} 
  \vspace{-0.4cm}
  \caption{
  Comparison of noise variance against signal intensity. The bottom grayscale strip represents increasing brightness. Naive sRGB-domain injection (orange) shows intensity-invariant variance, inconsistent with real sensors. In contrast, our PDS noise (red), generated via RAW-domain Poisson-Gaussian injection and ISP reprocessing, exhibits clear signal-dependent variance. This underscores the necessity of RAW-domain physics-based noise modeling for realistic low-light degradations.
  }
  \vspace{-0.3cm}
  \label{fig:noise_variance}
\end{figure}

\noindent \textbf{Stage II: Physical Noise Attack in the RAW Domain.}
In the linear RAW domain, sensor noise exhibits a fundamentally \emph{heteroscedastic} distribution that is intrinsically coupled with signal intensity. Unlike sRGB-space noise injection, where noise is typically assumed to be signal-independent, real imaging sensors produce noise whose variance grows with photon arrival rate.
To faithfully model this behavior, we adopt a Poisson--Gaussian noise formulation that explicitly accounts for both photon shot noise and electronic read noise.
We define an adversarial noise vector as $\mathcal{P}_{att} = \{k, \sigma\},$
where $k$ controls the signal-dependent Poisson component and $\sigma$ denotes the signal-independent Gaussian read noise. In practice, these parameters are randomly sampled in logarithmic space to cover a wide range of realistic imaging conditions.
Specifically, following real sensor statistics, we sample $k \in [10^{-3}, 10^{-2}]$ and $\sigma \in [10^{-4}, 5 \times 10^{-3}]$,
which is consistent with the noise parameter ranges used in our implementation.
The attacked RAW measurement is synthesized as
\begin{equation}
\hat{I}_{RAW} \sim \mathcal{N}
\left(
I_{RAW}, \; k \cdot I_{RAW} + \sigma^2
\right).
\end{equation}
Fig.~\ref{fig:noise_variance} provides a quantitative comparison between naive sRGB-domain noise injection and our RAW-domain PDS strategy.
When noise is directly injected in the sRGB domain, the variance remains nearly constant across different brightness levels, failing to capture the intensity-sensitive nature of real noise.
In contrast, PDS produces a clear intensity-dependent variance profile after forward ISP reprocessing, where noise sensitivity varies significantly from dark to bright regions.
This physics-based noise attack exposes failure modes that are difficult to simulate with additive sRGB noise and serves as a strong adversarial signal for training noise-aware enhancement networks.

\noindent \textbf{Stage III: ISP Reprocessing and \textbf{Physics-based Degraded Sample} Generation.}
To obtain the final physics-based degraded sample in the sRGB domain, the corrupted RAW signal undergoes a forward ISP simulation. Specifically, the Bayer-patterned signal is first demosaicked using a differentiable bilinear interpolation scheme, which introduces spatially correlated noise. The resulting linear RGB image is then processed by forward white balance, color correction using the sampled CCM, and gamma compression:
\begin{equation}
I_{att} =
\Gamma \left(
\mathbf{M}_{ccm} \cdot
\left(
\mathcal{D}(\hat{I}_{RAW}) \odot \mathbf{g}_{wb}
\right)
\right),
\end{equation}
where $I_{att}$ denotes the generated physics-based degraded image in the sRGB domain, 
$\hat{I}_{RAW}$ is the noise-corrupted RAW signal obtained after physical noise injection,
$\mathcal{D}(\cdot)$ represents the demosaicking operator,
$\mathbf{g}_{wb}$ denotes the channel-wise white balance gains,
$\odot$ indicates channel-wise multiplication,
$\mathbf{M}_{ccm}$ is the camera color correction matrix that maps camera RGB to sRGB,
and $\Gamma(\cdot)$ denotes the gamma compression function used in the forward ISP.
The generated $I_{att}$ serves as a physically faithful degraded sample with explicit ground-truth noise parameters $\mathcal{P}_{att}$.

\subsection{Noise Predictor}

In the defense pipeline, we design a lightweight Noise Predictor. This module acts as a \emph{sentinel} within the defense framework, aiming to explicitly estimate the physical noise parameters $\mathcal{P}_{pre}$ of a low-light input image $I_{input}$.
As shown in Fig.~\ref{fig:overall}, the Noise Predictor is composed of three $3\times3$ convolutional layers, each followed by a LeakyReLU activation. Such a shallow architecture is sufficient to capture high-frequency noise patterns while avoiding overfitting to semantic image content. A global average pooling layer is then employed to aggregate spatial features into a compact, content-independent embedding. Finally, a multi-layer perceptron (MLP) maps this embedding to the noise parameter space, followed by a Sigmoid activation that normalizes the predicted noise parameters $\tilde{k}$ and $\tilde{\sigma}$ to the range $[0,1]$.

\noindent \textbf{Optimization of Noise Predictor: Dual-Domain Self-Supervised Loss.}
The accuracy of the Noise Predictor's prediction directly determines the performance ceiling of the entire defense system, as inaccurate estimates may mislead subsequent adaptive modulation modules based on DA-MoE and SFT layers. We therefore optimize the Noise Predictor using a Dual-Domain Self-Supervised Loss strategy.

On the attack side, PDS provides physics-based degraded samples $I_{att}$ together with known noise attack parameters $\mathcal{P}_{att} = \{k, \sigma\}$. As illustrated in the normal-light domain branch of the Dual-Domain Self-Supervised Loss module in Fig.~\ref{fig:overall}, we first optimize the Noise Predictor by minimizing the mean squared error (MSE):
\begin{equation}
\mathcal{L}^{\text{normal}}_{\text{np}}
=\left\|\mathcal{NP}(I_{att}) - \mathcal{P}_{att}\right\|_2^2,
\label{eq:np_normal}
\end{equation}
where $\mathcal{NP}(\cdot)$ denotes the Noise Predictor. This self-supervised objective encourages the predicted noise parameters to accurately recover the injected physical noise attack parameters.

However, PDS performs noise attacks on normal-light images, whereas our ultimate goal is to predict the noise parameters of real low-light inputs $I_{input}$, denoted as $\mathcal{NP}(I_{input}) = \mathcal{P}_{pre}$, where $\mathcal{P}_{pre} = (\tilde{k}, \tilde{\sigma})$. The severe domain discrepancy makes it insufficient to optimize the accuracy of low-light noise prediction using Eq.~\ref{eq:np_normal} alone. Moreover, real low-light images are inherently corrupted by unknown noise, which renders direct supervision on absolute noise levels infeasible. 

To address these challenges, we introduce a self-perturbation-based additive noise prediction strategy that enforces relative physical consistency on unlabeled low-light images. Specifically, a low-light input image $I_{input}$ already contains unknown noise $N$ and thus should not be treated as a clean observation. We therefore apply an additional noise attack $\mathcal{P}_{perturb}=\{k_{per},\sigma_{per}\}$ to $I_{input}$ using PDS, yielding a perturbed degraded sample $I_{perturb}$. This process simulates an extra noise perturbation applied on top of the unknown noise already present in the low-light image. Notably, the injected perturbation is mainly applied to the Gaussian read-noise component, while the Poisson shot-noise component is set to be extremely small or kept unchanged. This design is motivated by the fact that read noise is signal-independent and can be injected in a physically well-defined manner, whereas introducing additional shot noise would entangle noise variation with unknown scene intensity, leading to ambiguous supervision. Consequently, we enforce \emph{shot-noise consistency} between $I_{input}$ and $I_{perturb}$, i.e., the predicted Poisson parameter should remain unchanged:
\begin{equation}
k_{per}\approx 0 \quad \Rightarrow \quad k(I_{perturb}) \approx k(I_{input})=\tilde{k}.
\end{equation}

Since independent Gaussian noise sources exhibit additive variance in the RAW domain, the effective read-noise level after perturbation should satisfy:
\begin{equation}
\delta=\sqrt{\tilde{\sigma}^2+\sigma_{per}^2}.
\end{equation}
Based on this physical constraint, we impose a consistency condition between the noise predictions of the original low-light input and its perturbed counterpart:
\begin{equation}
\mathcal{L}^{\text{low}}_{\text{np}}
=
\left\|
\mathcal{NP}(I_{perturb})
-
\left(\tilde{k}, \sqrt{\tilde{\sigma}^2 + \sigma_{per}^2}\right)
\right\|_2^2.
\end{equation}
This loss enforces that the noise parameters predicted for the more severely corrupted sample $I_{perturb}$ correspond to the baseline prediction $\mathcal{P}_{pre}$ augmented by the known read-noise increment, while keeping the shot-noise parameter consistent to avoid ill-posed supervision and degenerate solutions.

By jointly leveraging explicit supervision in the normal-light domain and physically constrained consistency learning in the low-light domain, the Noise Predictor can be effectively aligned across domains:
\begin{equation}
\mathcal{L}_{\text{consist}}
=\mathcal{L}^{\text{normal}}_{\text{np}}+\mathcal{L}^{\text{low}}_{\text{np}}.
\label{eq:consist}
\end{equation}
This training strategy encourages the Noise Predictor to follow physically meaningful noise evolution rules rather than exploiting spurious correlations in the data. Consequently, the Noise Predictor provides reliable and interpretable noise parameter priors for subsequent DA-MoE and SFT layers.

\subsection{\textbf{Degradation-Aware Mixture-of-Experts (DA-MoE)}}
In real-world low-light imaging, noise is inherently heteroscedastic, arising from the combination of signal-independent read noise (Gaussian noise) and signal-dependent photon shot noise (Poisson noise). Conventional convolutional neural networks (CNNs), due to their fixed weights, struggle to achieve an optimal balance between noise suppression and detail preservation across these distinct noise modalities. To address this limitation, we propose a DA-MoE module, as shown in Fig.~\ref{fig:overall} (upper-right). The DA-MoE is integrated into existing architectures in a plug-and-play manner and functions as a dynamic router that fuses two specialized experts conditioned on the predicted noise parameters $\mathcal{P}_{pre}$.

Specifically, the DA-MoE consists of two complementary processing branches:

\noindent \textbf{Gaussian Noise Expert ($E_g$):} This expert is designed to handle signal-independent image degradation, i.e., read noise. It adopts standard residual blocks with spatially shared weights and focuses on suppressing global additive disturbances independent of image content.

\noindent \textbf{Poisson Noise Expert ($E_p$):} This expert targets signal-dependent degradation caused by photon shot noise. Since the variance of shot noise varies with pixel intensity, static convolutions are suboptimal. Therefore, this branch incorporates a SFT layer~\cite{wang2018sftgan} to enable pixel-wise adaptive modulation, allowing stronger denoising in regions where shot noise is more severe.

A lightweight gating network then maps the predicted noise vector $\mathcal{P}_{pre}$ to fusion weights via a Softmax function:
\[
[w_g, w_p] = \text{Softmax}(\text{MLP}(\mathcal{P}_{pre})).
\]
The final output feature is obtained as a weighted combination of the expert outputs:
\begin{equation}
F_{out} = w_g \cdot E_g(F_{in}) + w_p \cdot E_p(F_{in}),
\end{equation}
where $F_{in}$ and $F_{out}$ denote intermediate feature representations within the network.
Notably, as image features are progressively downsampled and averaged by the encoder, neighboring pixels are effectively merged. Due to the Central Limit Theorem (CLT), independent Poisson-distributed variables, when summed or averaged, asymptotically converge to a Gaussian distribution. Consequently, at the low-resolution bottleneck, signal-dependent characteristics diminish, and the noise distribution becomes approximately Gaussian. Thus, only the Gaussian expert is employed at this stage.

\noindent \textbf{Spatial Feature Transform (SFT) Layer.}
Standard convolutional layers rely on spatially shared weights and implicitly assume spatially homogeneous degradation (i.e., translation invariance). However, low-light images are affected by signal-dependent shot noise, whose variance varies significantly with pixel intensity. As a result, bright and dark regions require different denoising strengths. Applying static convolutional kernels uniformly across the image may lead to insufficient denoising in dark regions or over-smoothing in textured areas.

To overcome this limitation, we adopt the SFT layer~\cite{wang2018sftgan}. The core idea of SFT is to introduce a physics-guided affine modulation mechanism that acts as a dynamic gate. Conditioned on the noise parameters $\mathcal{P}_{pre}$, SFT performs pixel-wise modulation of feature distributions, enabling the network to adaptively adjust its behavior according to local noise strength. Specifically, SFT learns spatially varying scale ($\mathbf{\gamma}$) and shift ($\mathbf{\beta}$) parameters, which effectively bridge the gap between global physical priors and local spatial enhancement.

As shown in Fig.~\ref{fig:overall}, the predicted noise parameters $\mathcal{P}_{pre}$ are first mapped to a high-dimensional latent vector via an MLP. This vector is then concatenated with the input feature map $F'_{in}$ and fed into convolutional and activation layers to extract spatially varying control signals. The resulting features are split into two branches: a $1\times1$ convolution produces the scale map $\mathbf{\gamma}$, while another $1\times1$ convolution generates the shift map $\mathbf{\beta}$. Finally, the input features are modulated through an affine transformation:
\begin{equation}
F^{'}_{out} = \text{SFT}(F^{'}_{in}\mid\mathcal{P}_{pre})
= (1 + \mathbf{\gamma}) \odot F^{'}_{in} + \mathbf{\beta},
\end{equation}
where $\odot$ denotes element-wise multiplication.

\subsection{\textbf{Adaptive Metric Defense (AMD)}}
\label{subsubsec:dynamic_defense}

Although the DA-MoE performs adaptive image enhancement in the spatial domain through structural modulation, structural constraints alone cannot guarantee the complete removal of all artifacts. Conventional pixel-wise loss functions, such as $\ell_1$ or MSE, treat image enhancement as a uniform regression problem and ignore the varying difficulty induced by different degradation levels. Since they do not penalize the perceptual proximity between residual artifacts and noisy inputs, such losses often lead to over-smoothed results. To address these limitations, we introduce an AMD. As shown in Fig.~\ref{fig:overall} (lower-right), we construct a triplet in the VGG feature space, comprising the enhanced output $I_{output}$, the clean reference image $I_{normal}$, and the physics-based degraded sample $I_{att}$ generated by PDS. The key observation is that standard triplet or contrastive losses enforce a fixed discriminative margin. However, physics-based degraded samples with weaker noise perturbations lie closer to the clean decision boundary and therefore require a stronger repulsive force.

To this end, we design a dynamic margin $m$ that is explicitly modulated by the noise attack parameters $\mathcal{P}_{att} = \{k, \sigma\}$ derived from the PDS stream. We quantify the overall degradation magnitude using the $\ell_2$ norm of the noise parameter vector and map it to the metric space with a scaling factor:
\begin{equation}
m(\mathcal{P}_{att}) = \eta \, \lVert \mathcal{P}_{att} \rVert_2,
\end{equation}
where $\eta$ is a learnable or predefined scalar that aligns the physical noise magnitude with the scale of VGG feature distances.
Let $\phi_i(\cdot)$ denote the feature representation extracted from the $i$-th layer of a pretrained VGG-19~\cite{simonyan2014very}. The perceptual distance between two images $X$ and $Y$ at layer $i$ is defined as
\begin{equation}
d_i(X, Y) = \lVert \phi_i(X) - \phi_i(Y) \rVert_1.
\end{equation}

Given the enhanced image $I_{output}$ as the \emph{anchor}, the clean image $I_{normal}$ as the \emph{positive}, and the physics-based degraded sample $I_{att}$ as the \emph{negative}, our AMD is formulated as
\begin{equation}
\mathcal{L}_{\text{metric}}
=
\sum_i
\frac{
d_i(I_{output}, I_{normal})
}{
\max \bigl( d_i(I_{output}, I_{att}) - m(\mathcal{P}_{att}),\, 0 \bigr)
+ \epsilon
},
\label{eq:metric}
\end{equation}
where $i=[1, 6, 11, 20, 29]$ represent different layers of the VGG-19, and $\epsilon = 10^{-7}$ is a small constant for numerical stability.

This formulation establishes a noise-adaptive exclusion zone around the anchor in the perceptual feature space. When the negative degraded sample $I_{att}$ is insufficiently separated from the anchor (i.e., $d_i(I_{output}, I_{att}) < m$), the denominator becomes small, resulting in a large loss that strongly pushes the enhanced output away from the degraded direction. Conversely, for heavily degraded samples, a larger margin is permitted, leading to a smoother optimization landscape. Importantly, this AMD is implemented in the VGG-19 feature space rather than the pixel domain. Pixel-wise distances correlate poorly with human perception: small spatial shifts can cause large MSE values without perceptual significance. In contrast, VGG features encode semantic and textural information, allowing us to explicitly enforce perceptual separability between enhanced outputs, clean references, and physics-based degraded inputs.

\subsection{Optimization Objective}
\label{subsec:loss}

The overall loss function consists of three components: the reconstruction loss, the Dual-Domain Self-Supervised Loss ($\mathcal{L}_{consist}$, Eq.~\ref{eq:consist}), and the AMD ($\mathcal{L}_{metric}$, Eq.~\ref{eq:metric}).
The final optimization objective is formulated as:
\begin{equation}
\mathcal{L}_{total}
=
\mathcal{L}_{rec}
+
\lambda_{con}\mathcal{L}_{consist}
+
\lambda_{met}\mathcal{L}_{metric}.
\end{equation}
Based on empirical tuning, we set the balancing hyperparameters to $\lambda_{con}=0.5$ and $\lambda_{met}=0.01$ in all experiments.

\noindent \textbf{Reconstruction Loss ($\mathcal{L}_{rec}$).}  
The reconstruction loss ensures pixel-wise fidelity between the enhanced output and the ground-truth clean image. \emph{We keep the reconstruction loss unchanged to ensure a fair comparison with existing methods.}

\begin{table*}[!t]
    \renewcommand{\arraystretch}{1.0}
    \setlength{\tabcolsep}{3pt}
    \centering
    \caption{Results on LOL-Blur and LOL-v1 Datasets. Our plug-in almost improves PSNR/SSIM and reduces LPIPS/FID across all baselines, with particularly large gains on LPIPS and FID.}
    \vspace{-0.2cm}
    \begin{tabular}{l|c|cccc|cccc}
        \toprule
        \multirow{2}{*}{\centering \textbf{Method}} & \multirow{2}{*}{\centering \textbf{Venue}} & \multicolumn{4}{c|}{\textbf{LOL-Blur}} & \multicolumn{4}{c}{\textbf{LOL-v1}} \\
        \cline{3-10}
        & & PSNR $\uparrow$ & SSIM $\uparrow$ & LPIPS $\downarrow$ & FID $\downarrow$ & PSNR $\uparrow$ & SSIM $\uparrow$ & LPIPS $\downarrow$ & FID $\downarrow$ \\
        \midrule

        LEDNet~\cite{zhou2022lednet} &  ECCV 22 & 25.79 & 0.8500 & 0.1670 & 17.89 & 23.32 & 0.9029 & 0.0972 & 40.38 \\
        \rowcolor{gray!9}
        LEDNet+ & Ours
        & 26.40{\textcolor{red}{\scriptsize (+0.61)}}
        & 0.8664{\textcolor{red}{\scriptsize (+0.0164)}}
        & 0.1282{\textcolor{red}{\scriptsize (-0.0388)}}
        & 13.95{\textcolor{red}{\scriptsize (-3.94)}}
        & 23.37{\textcolor{red}{\scriptsize (+0.05)}}
        & 0.9014{\textcolor{green}{\scriptsize (-0.0015)}}
        & 0.0868{\textcolor{red}{\scriptsize (-0.0104)}}
        & 34.75{\textcolor{red}{\scriptsize (-5.63)}} \\ \hline \hline

        Restormer~\cite{zamir2022restormer} &  CVPR 22 & 25.21 & 0.8367 & 0.2190 & 25.19 & 23.31 & 0.8865 & 0.1298 & 62.70 \\
        \rowcolor{gray!9}
        Restormer+ & Ours
        & 25.83{\textcolor{red}{\scriptsize (+0.62)}}
        & 0.8655{\textcolor{red}{\scriptsize (+0.0288)}}
        & 0.1214{\textcolor{red}{\scriptsize (-0.0976)}}
        & 13.81{\textcolor{red}{\scriptsize (-11.38)}}
        & 23.45{\textcolor{red}{\scriptsize (+0.14)}}
        & 0.9033{\textcolor{red}{\scriptsize (+0.0168)}}
        & 0.0874{\textcolor{red}{\scriptsize (-0.0424)}}
        & 35.32{\textcolor{red}{\scriptsize (-27.38)}} \\ \hline \hline

        FourLLIE~\cite{wang2023fourllie} &  MM 23 & 22.48 & 0.7289 & 0.2596 & 26.85 & 20.06 & 0.8401 & 0.2409 & 88.17 \\
        \rowcolor{gray!9}
        FourLLIE+ & Ours
        & 24.59{\textcolor{red}{\scriptsize (+2.11)}}
        & 0.7862{\textcolor{red}{\scriptsize (+0.0573)}}
        & 0.1840{\textcolor{red}{\scriptsize (-0.0756)}}
        & 19.56{\textcolor{red}{\scriptsize (-7.29)}}
        & 21.34{\textcolor{red}{\scriptsize (+1.28)}}
        & 0.8815{\textcolor{red}{\scriptsize (+0.0414)}}
        & 0.1524{\textcolor{red}{\scriptsize (-0.0885)}}
        & 76.87{\textcolor{red}{\scriptsize (-11.30)}} \\ \hline \hline

        Retinexformer~\cite{retinexformer} &  ICCV 23 & 25.61 & 0.8274 & 0.2451 & 27.84 & 23.49 & 0.8882 & 0.1394 & 74.35 \\
        \rowcolor{gray!9}
        Retinexformer+ & Ours
        & 26.49{\textcolor{red}{\scriptsize (+0.88)}}
        & 0.8641{\textcolor{red}{\scriptsize (+0.0367)}}
        & 0.1220{\textcolor{red}{\scriptsize (-0.1231)}}
        & 13.60{\textcolor{red}{\scriptsize (-14.24)}}
        & 23.95{\textcolor{red}{\scriptsize (+0.46)}}
        & 0.9083{\textcolor{red}{\scriptsize (+0.0201)}}
        & 0.0933{\textcolor{red}{\scriptsize (-0.0461)}}
        & 42.92{\textcolor{red}{\scriptsize (-31.43)}} \\ \hline \hline

        Wave-Mamba~\cite{zou2024wave} &  MM 24 & 24.73 & 0.8240 & 0.2571 & 27.90 & 22.79 & 0.8987 & 0.1333 & 63.98 \\
        \rowcolor{gray!9}
        Wave-Mamba+ & Ours
        & 24.91{\textcolor{red}{\scriptsize (+0.18)}}
        & 0.8288{\textcolor{red}{\scriptsize (+0.0048)}}
        & 0.1988{\textcolor{red}{\scriptsize (-0.0583)}}
        & 22.94{\textcolor{red}{\scriptsize (-4.96)}}
        & 22.86{\textcolor{red}{\scriptsize (+0.07)}}
        & 0.8927{\textcolor{green}{\scriptsize (-0.0060)}}
        & 0.1127{\textcolor{red}{\scriptsize (-0.0206)}}
        & 48.57{\textcolor{red}{\scriptsize (-15.41)}} \\ \hline \hline

        RetinexMamba~\cite{bai2024retinexmamba} &  ICNIP 24 & 25.53 & 0.8263 & 0.2551 & 29.98 & 22.87 & 0.8809 & 0.1414 & 71.56 \\
        \rowcolor{gray!9}
        RetinexMamba+ & Ours
        & 26.56{\textcolor{red}{\scriptsize (+1.03)}}
        & 0.8746{\textcolor{red}{\scriptsize (+0.0483)}}
        & 0.1081{\textcolor{red}{\scriptsize (-0.1470)}}
        & 11.92{\textcolor{red}{\scriptsize (-18.06)}}
        & 23.78{\textcolor{red}{\scriptsize (+0.91)}}
        & 0.9057{\textcolor{red}{\scriptsize (+0.0248)}}
        & 0.0949{\textcolor{red}{\scriptsize (-0.0465)}}
        & 43.01{\textcolor{red}{\scriptsize (-28.55)}} \\ \hline \hline

        DMFourLLIE~\cite{zhang2024dmfourllie} &  MM 24 & 21.20 & 0.8247 & 0.2386 & 24.71 & 22.98 & 0.8860 & 0.1270 & 60.83 \\
        \rowcolor{gray!9}
        DMFourLLIE+ & Ours
        & 24.95{\textcolor{red}{\scriptsize (+3.75)}}
        & 0.8683{\textcolor{red}{\scriptsize (+0.0436)}}
        & 0.1792{\textcolor{red}{\scriptsize (-0.0594)}}
        & 18.52{\textcolor{red}{\scriptsize (-6.19)}}
        & 24.03{\textcolor{red}{\scriptsize (+1.05)}}
        & 0.9081{\textcolor{red}{\scriptsize (+0.0221)}}
        & 0.0930{\textcolor{red}{\scriptsize (-0.0340)}}
        & 40.82{\textcolor{red}{\scriptsize (-20.01)}} \\ \hline \hline

        CWNet~\cite{zhang2025cwnet} &  ICCV 25 & 25.59 & 0.8542 & 0.1883 & 19.95 & 23.60 & 0.9023 & 0.1230 & 57.53 \\
        \rowcolor{gray!9}
        CWNet+ & Ours
        & 26.07{\textcolor{red}{\scriptsize (+0.48)}}
        & 0.8753{\textcolor{red}{\scriptsize (+0.0211)}}
        & 0.1252{\textcolor{red}{\scriptsize (-0.0631)}}
        & 14.13{\textcolor{red}{\scriptsize (-5.82)}}
        & 23.57{\textcolor{green}{\scriptsize (-0.03)}}
        & 0.9037{\textcolor{red}{\scriptsize (+0.0014)}}
        & 0.0901{\textcolor{red}{\scriptsize (-0.0329)}}
        & 39.05{\textcolor{red}{\scriptsize (-18.48)}} \\ \hline \hline

        FDN~\cite{tu2025fourier} &  TIP 25 & 28.19 & 0.9180 & 0.0820 & 8.51 & 23.99 & 0.9100 & 0.0830 & 32.79 \\
        \rowcolor{gray!9}
        FDN+ & Ours
        & 28.41{\textcolor{red}{\scriptsize (+0.22)}}
        & 0.9201{\textcolor{red}{\scriptsize (+0.0021)}}
        & 0.0784{\textcolor{red}{\scriptsize (-0.0036)}}
        & 7.57{\textcolor{red}{\scriptsize (-0.94)}}
        & 24.31{\textcolor{red}{\scriptsize (+0.32)}}
        & 0.9135{\textcolor{red}{\scriptsize (+0.0035)}}
        & 0.0784{\textcolor{red}{\scriptsize (-0.0046)}}
        & 29.52{\textcolor{red}{\scriptsize (-3.27)}} \\
        \bottomrule
    \end{tabular}
     \vspace{-0.2cm}
    \label{tab:com_lol-blur_lol-v1}
\end{table*}

\begin{figure*}[!t]
    \centering
    \includegraphics[width=1\linewidth]{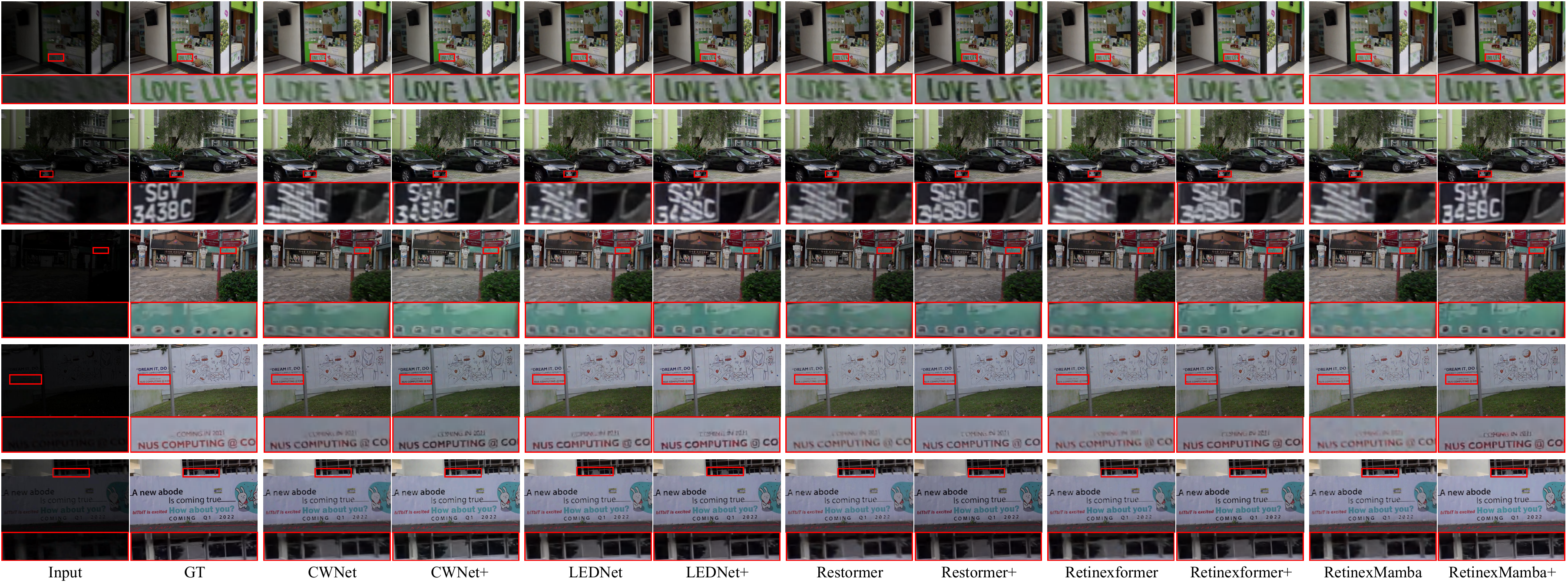}
     \vspace{-0.8cm}
    \caption{Visual comparison on the LOL-Blur dataset. The first two columns are the input and ground truth. For each method, we show the baseline result and its enhanced variant, where ``+'' indicates integrating our proposed adversarial framework. Zoom in for better inspection of noise suppression.}
     \vspace{-0.3cm}
    \label{fig:vis_lol_blur}
\end{figure*}

\begin{figure*}[!t]
    \centering
    \includegraphics[width=0.95\linewidth]{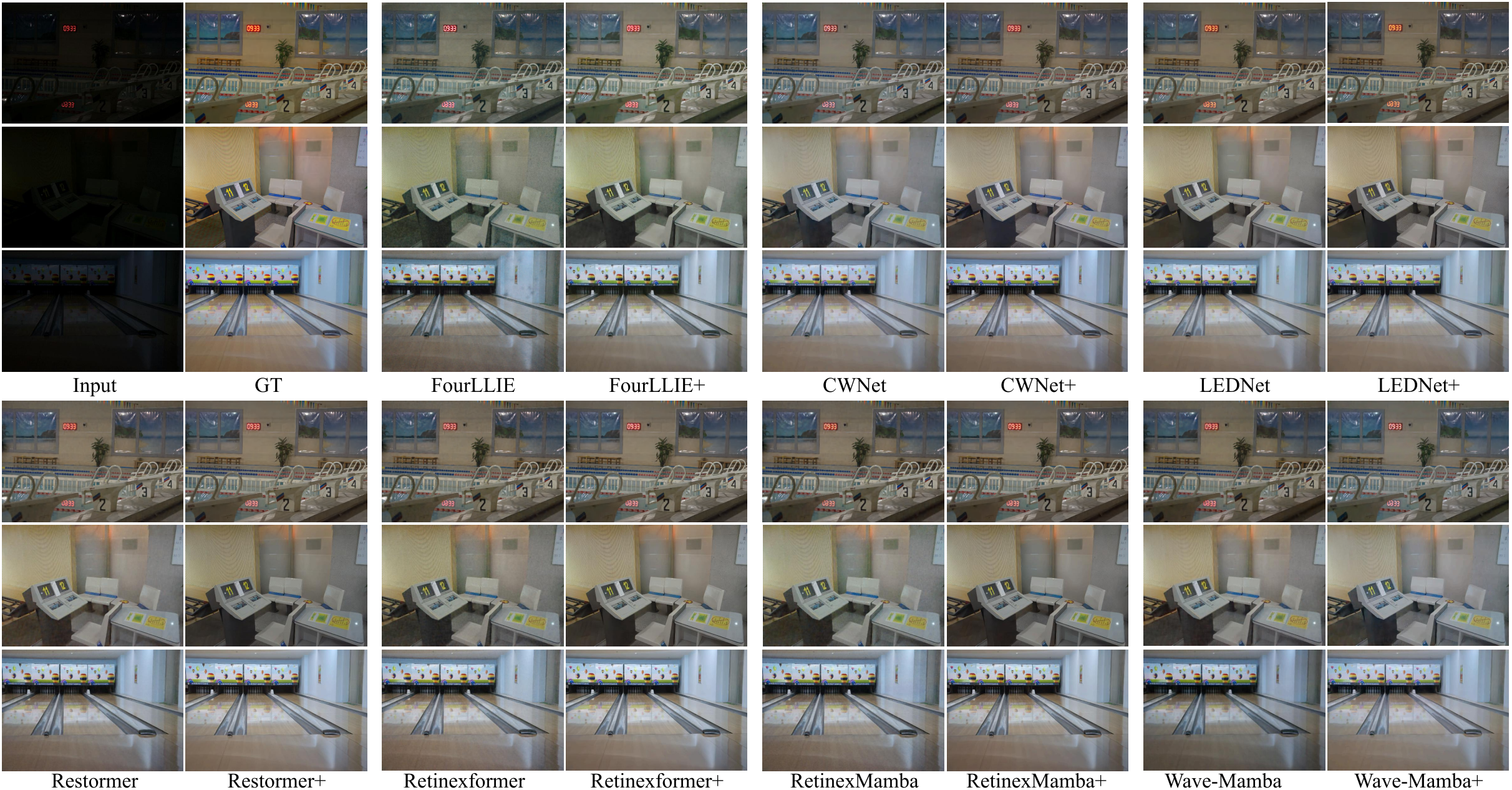}
     \vspace{-0.3cm}
    \caption{Visual comparisons with state-of-the-art methods on LOL-v1 dataset. Zoom in for better detail recovery.}
     \vspace{-0.6cm}
    \label{fig:vis_lol_v1}
\end{figure*}

\begin{table}[!t]
    \renewcommand{\arraystretch}{1.0}
    \setlength{\tabcolsep}{3pt}
    \centering
    \caption{Results on the Real-LOL-Blur dataset using models pretrained on LOL-Blur. }
     \vspace{-0.2cm}
    \begin{tabular}{l|ccc}
        \toprule
        \multirow{2}{*}{\centering \textbf{Method}} & \multicolumn{3}{c}{\textbf{Real-LOL-Blur}} \\
        \cline{2-4}
        & NIQE $\downarrow$ & BRISQUE $\downarrow$ & PI $\downarrow$ \\
        \midrule

        LEDNet~\cite{zhou2022lednet} & 5.169 & 42.801 & 5.030 \\
        \rowcolor{gray!9}
        LEDNet+ & 4.506{\textcolor{red}{\scriptsize (-0.663)}} & 35.609{\textcolor{red}{\scriptsize (-7.192)}} & 4.339{\textcolor{red}{\scriptsize (-0.691)}} \\ \hline \hline

        Restormer~\cite{zamir2022restormer} & 5.457 & 36.682 & 5.488 \\
        \rowcolor{gray!9}
        Restormer+ & 4.154{\textcolor{red}{\scriptsize (-1.303)}} & 25.911{\textcolor{red}{\scriptsize (-10.771)}} & 3.874{\textcolor{red}{\scriptsize (-1.614)}} \\ \hline \hline

        FourLLIE~\cite{wang2023fourllie} & 5.983 & 43.892 & 6.214 \\
        \rowcolor{gray!9}
        FourLLIE+ & 5.821{\textcolor{red}{\scriptsize (-0.162)}} & 42.962{\textcolor{red}{\scriptsize (-0.930)}} & 5.974{\textcolor{red}{\scriptsize (-0.240)}} \\ \hline \hline

        Retinexformer~\cite{retinexformer} & 5.860 & 38.286 & 5.873 \\
        \rowcolor{gray!9}
        Retinexformer+ & 4.408{\textcolor{red}{\scriptsize (-1.452)}} & 31.392{\textcolor{red}{\scriptsize (-6.894)}} & 4.129{\textcolor{red}{\scriptsize (-1.744)}} \\ \hline \hline

        Wave-Mamba~\cite{zou2024wave} & 5.719 & 42.975 & 5.775 \\
        \rowcolor{gray!9}
        Wave-Mamba+ & 5.314{\textcolor{red}{\scriptsize (-0.405)}} & 38.474{\textcolor{red}{\scriptsize (-4.501)}} & 5.291{\textcolor{red}{\scriptsize (-0.484)}} \\ \hline \hline

        RetinexMamba~\cite{bai2024retinexmamba} & 5.802 & 37.393 & 5.808 \\
        \rowcolor{gray!9}
        RetinexMamba+ & 4.193{\textcolor{red}{\scriptsize (-1.609)}} & 25.387{\textcolor{red}{\scriptsize (-12.006)}} & 3.855{\textcolor{red}{\scriptsize (-1.953)}} \\ \hline \hline

        DMFourLLIE~\cite{zhang2024dmfourllie} & 5.796 & 42.173 & 5.832 \\
        \rowcolor{gray!9}
        DMFourLLIE+ & 5.298{\textcolor{red}{\scriptsize (-0.498)}} & 37.482{\textcolor{red}{\scriptsize (-4.691)}} & 5.397{\textcolor{red}{\scriptsize (-0.435)}} \\ \hline \hline

        CWNet~\cite{zhang2025cwnet} & 5.175 & 37.335 & 5.155 \\
        \rowcolor{gray!9}
        CWNet+ & 4.193{\textcolor{red}{\scriptsize (-0.982)}} & 25.387{\textcolor{red}{\scriptsize (-11.948)}} & 3.855{\textcolor{red}{\scriptsize (-1.300)}} \\ \hline \hline

        FDN~\cite{tu2025fourier} & 4.133 & 15.95 & 3.670 \\
        \rowcolor{gray!9}
        FDN+ & 3.875{\textcolor{red}{\scriptsize (-0.258)}} & 14.27{\textcolor{red}{\scriptsize (-1.68)}} & 3.549{\textcolor{red}{\scriptsize (-0.121)}} \\
        \bottomrule
    \end{tabular}
     \vspace{-0.4cm}
    \label{tab:lol-blur-real-lol-blur}
\end{table}

\section{Experiments}
\label{sec:experiments}

\subsection{Datasets and Implementation Details}
We train and evaluate on three standard real-world low-light benchmarks spanning degradations from mild low light to challenging mixtures of noise and blur.
\textbf{LOL-Blur}~\cite{zhou2022lednet} targets joint low-light and motion blur, providing 12{,}000 paired images (10{,}200 train / 1{,}800 test) with low-light blurry inputs and normal-light sharp ground truth.
\textbf{Real-LOL-Blur}~\cite{zhou2022lednet} contains 1{,}354 unpaired real low-light blurry images without ground truth for in-the-wild qualitative evaluation.
\textbf{LOL-v1}~\cite{lol} is a classic paired LLIE benchmark with 500 real image pairs; we follow the official split (485 train / 15 test).

Our framework is implemented in PyTorch and trained on two NVIDIA RTX 4090 GPUs. 
To ensure a rigorous and fair evaluation, we strictly align our training protocol with the official configurations of all baseline methods. Specifically, the learning rate, input patch size, data augmentation strategies (e.g., random flip and rotation), batch size, optimizer settings, and total training iterations are kept identical to those reported in their respective official implementations. 
For methods requiring specific loss weight configurations, we adopt the optimal settings provided by the original authors. This ensures that any performance gain is attributed to our architectural innovations rather than hyperparameter tuning.

\begin{table}[!t]
    \renewcommand{\arraystretch}{1.0}
    \setlength{\tabcolsep}{3pt}
    \centering
    \caption{Results on the Real-LOL-Blur dataset using models pretrained on LOL-v1.}
     \vspace{-0.2cm}
    \begin{tabular}{l|ccc}
        \toprule
        \multirow{2}{*}{\centering \textbf{Method}} & \multicolumn{3}{c}{\textbf{Real-LOL-Blur}} \\
        \cline{2-4}
        & NIQE $\downarrow$ & BRISQUE $\downarrow$ & PI $\downarrow$ \\
        \midrule

        LEDNet~\cite{zhou2022lednet} & 5.322 & 31.849 & 5.535 \\
        \rowcolor{gray!9}
        LEDNet+ & 5.280{\textcolor{red}{\scriptsize (-0.042)}} & 30.419{\textcolor{red}{\scriptsize (-1.430)}} & 5.486{\textcolor{red}{\scriptsize (-0.049)}} \\ \hline \hline

        Restormer~\cite{zamir2022restormer} & 5.287 & 34.544 & 5.461 \\
        \rowcolor{gray!9}
        Restormer+ & 4.925{\textcolor{red}{\scriptsize (-0.362)}} & 32.913{\textcolor{red}{\scriptsize (-1.631)}} & 4.898{\textcolor{red}{\scriptsize (-0.563)}} \\ \hline \hline

        FourLLIE~\cite{wang2023fourllie} & 5.796 & 41.620 & 5.791 \\
        \rowcolor{gray!9}
        FourLLIE+ & 5.655{\textcolor{red}{\scriptsize (-0.141)}} & 38.271{\textcolor{red}{\scriptsize (-3.349)}} & 5.547{\textcolor{red}{\scriptsize (-0.244)}} \\ \hline \hline

        Retinexformer~\cite{retinexformer} & 5.057 & 25.302 & 5.214 \\
        \rowcolor{gray!9}
        Retinexformer+ & 4.828{\textcolor{red}{\scriptsize (-0.229)}} & 26.453{\textcolor{green}{\scriptsize (+1.151)}} & 4.893{\textcolor{red}{\scriptsize (-0.321)}} \\ \hline \hline

        Wave-Mamba~\cite{zou2024wave} & 5.237 & 37.649 & 5.507 \\
        \rowcolor{gray!9}
        Wave-Mamba+ & 5.209{\textcolor{red}{\scriptsize (-0.028)}} & 34.631{\textcolor{red}{\scriptsize (-3.018)}} & 5.408{\textcolor{red}{\scriptsize (-0.099)}} \\ \hline \hline

        RetinexMamba~\cite{bai2024retinexmamba} & 4.956 & 32.837 & 5.167 \\
        \rowcolor{gray!9}
        RetinexMamba+ & 4.792{\textcolor{red}{\scriptsize (-0.164)}} & 30.760{\textcolor{red}{\scriptsize (-2.077)}} & 4.909{\textcolor{red}{\scriptsize (-0.258)}} \\ \hline \hline

        DMFourLLIE~\cite{zhang2024dmfourllie} & 5.585 & 38.426 & 5.631 \\
        \rowcolor{gray!9}
        DMFourLLIE+ & 5.472{\textcolor{red}{\scriptsize (-0.113)}} & 35.831{\textcolor{red}{\scriptsize (-2.595)}} & 5.500{\textcolor{red}{\scriptsize (-0.131)}} \\ \hline \hline

        CWNet~\cite{zhang2025cwnet} & 5.413 & 41.140 & 5.843 \\
        \rowcolor{gray!9}
        CWNet+ & 4.193{\textcolor{red}{\scriptsize (-1.220)}} & 25.387{\textcolor{red}{\scriptsize (-15.753)}} & 3.855{\textcolor{red}{\scriptsize (-1.988)}} \\ \hline \hline

        FDN~\cite{tu2025fourier} & 5.034 & 31.597 & 5.286 \\
        \rowcolor{gray!9}
        FDN+ & 4.275{\textcolor{red}{\scriptsize (-0.759)}} & 24.910{\textcolor{red}{\scriptsize (-6.687)}} & 4.084{\textcolor{red}{\scriptsize (-1.202)}} \\
        \bottomrule
    \end{tabular}
     \vspace{-0.4cm}
    \label{tab:lol-blur-real_v1}
\end{table}

\subsection{Comparison with State-of-the-Art Methods}
We compare our proposed method against a comprehensive set of state-of-the-art (SOTA) approaches, representing various lines of research and categorized by their core mechanisms.
For frequency-based approaches, we include FourLLIE~\cite{wang2023fourllie}, FDN~\cite{tu2025fourier} (a Fourier-based joint deblurring and low-light enhancement method), and DMFourLLIE~\cite{zhang2024dmfourllie} (a Fourier-based multimodal fusion framework). To reflect recent advances in State Space Models (SSMs), we also compare with Mamba-based architectures including Wave-Mamba~\cite{zou2024wave} and RetinexMamba~\cite{bai2024retinexmamba}. For CNN-based joint restoration under mixed degradations, we adopt LEDNet~\cite{zhou2022lednet}. Finally, for transformer-based methods, we evaluate against the specialized low-light transformer Retinexformer~\cite{cai2023retinexformer} as well as the general image restoration transformer Restormer~\cite{zamir2022restormer}, to demonstrate the effectiveness of our domain-specific design.

For the paired LOL-v1 and LOL-Blur datasets, we use PSNR and SSIM as full-reference metrics, together with LPIPS~\cite{zhang2018unreasonable} and FID~\cite{heusel2017gans} for perceptual quality evaluation. For Real-LOL-Blur, we adopt widely used no-reference image quality assessment metrics, including NIQE~\cite{mittal2012making}, BRISQUE~\cite{mittal2012no}, and PI~\cite{blau20182018}.

\subsubsection{Evaluation on LOL-Blur Dataset}
\label{subsec:exp_lol_blur}

\noindent\textbf{Quantitative Comparison.}
Tab.~\ref{tab:com_lol-blur_lol-v1} reports the results on LOL-Blur, where ``+'' denotes the corresponding baseline equipped with our proposed adversarial framework.
Our approach consistently yields improvements across all baselines on all four metrics: PSNR/SSIM increase, while LPIPS/FID decrease.
Notably, the gains are most evident on perceptual metrics, with substantial LPIPS reductions (up to $-0.1470$) and FID reductions (up to $-18.06$). This indicates that our method effectively suppresses unrealistic artifacts, producing outputs perceptually closer to the ground-truth distribution, and also improves distortion-oriented metrics.

\noindent\textbf{Qualitative Comparison.}
Fig.~\ref{fig:vis_lol_blur} presents representative visual comparisons on LOL-Blur.
Overall, compared with the original baselines, the ``+'' variants generate sharper structures with better noise suppression, improved detail recovery, and more faithful color reproduction.
Row 1: Our method enhances clarity and recovers fine textures without amplifying noise in dark regions.
Row 2: While several baselines introduce noticeable artifacts around high-frequency structures (e.g., edges and characters), our method effectively suppresses these and restores cleaner, more legible details.
Row 3: Our results exhibit clearer color rendition and richer details, coupled with reduced residual noise.
Row 4: Our method better reconstructs text-like structures and simultaneously controls noise interference over large, low-texture areas (e.g., the curtain/screen region).
Row 5: Our approach produces the most natural and clear enhancement, alleviating both blur and noise while avoiding over-enhancement.

\begin{table}[!t]
    \renewcommand{\arraystretch}{1.0}
    \setlength{\tabcolsep}{1.6pt}
    \centering
    \caption{Quantitative results on the GoPro and SID datasets.}
    \vspace{-0.2cm}
    \begin{tabular}{l|cc|cc}
        \toprule
        \multirow{2}{*}{\centering \textbf{Method}} & \multicolumn{2}{c|}{\textbf{GoPro}} & \multicolumn{2}{c}{\textbf{SID}} \\
        \cline{2-5}
        & PSNR $\uparrow$ & SSIM $\uparrow$ & PSNR $\uparrow$ & SSIM $\uparrow$ \\
        \midrule

        Retinexformer~\cite{retinexformer} & 29.14 & 0.8808 & 25.74 & 0.6951 \\
        \rowcolor{gray!9}
        Retinexformer+
        & 29.10{\textcolor{green}{\scriptsize (-0.04)}}
        & 0.8839{\textcolor{red}{\scriptsize (+0.0031)}}
        & 26.80{\textcolor{red}{\scriptsize (+1.06)}}
        & 0.7012{\textcolor{red}{\scriptsize (+0.0061)}} \\ \hline \hline

        Wave-Mamba~\cite{zou2024wave} & 26.94 & 0.8273 & 23.57 & 0.6740 \\
        \rowcolor{gray!9}
        Wave-Mamba+
        & 27.03{\textcolor{red}{\scriptsize (+0.09)}}
        & 0.8296{\textcolor{red}{\scriptsize (+0.0023)}}
        & 23.98{\textcolor{red}{\scriptsize (+0.41)}}
        & 0.6846{\textcolor{red}{\scriptsize (+0.0106)}} \\ \hline \hline

        CWNet~\cite{zhang2025cwnet} & 30.71 & 0.9104 & 25.82 & 0.7024 \\
        \rowcolor{gray!9}
        CWNet+
        & 31.08{\textcolor{red}{\scriptsize (+0.37)}}
        & 0.9123{\textcolor{red}{\scriptsize (+0.0019)}}
        & 26.14{\textcolor{red}{\scriptsize (+0.32)}}
        & 0.7052{\textcolor{red}{\scriptsize (+0.0028)}} \\ 
        \bottomrule
    \end{tabular}
    \vspace{-0.3cm}
    \label{tab:gopro_sid}
\end{table}

\subsubsection{Evaluation on LOL-v1 Dataset}
\label{subsec:exp_lol_v1}

\noindent\textbf{Quantitative Comparison.}
The LOL-v1 results are reported in Tab.~\ref{tab:com_lol-blur_lol-v1}.
Overall, integrating our framework improves performance for most baselines. We observe consistent improvements on perceptual metrics (LPIPS and FID), while PSNR shows a slight decrease for some backbones (e.g., CWNet). We attribute this to a common perception--fidelity trade-off in sRGB LLIE: our PDS-generated samples encourage the model to handle more realistic and entangled noise patterns, and the adaptive defense prioritizes noise suppression and perceptual naturalness.
However, PSNR is highly sensitive to pixel-wise deviations caused by nonlinear tone/color mapping and denoising-induced texture redistribution in the sRGB domain, which may not correlate well with visual quality under real low-light degradations.
Quantitatively, the PSNR drop is small compared to the perceptual gains.
For CWNet, PSNR changes from 23.60 to 23.57 ($-0.03$\,dB), while LPIPS improves from 0.1230 to 0.0901 and FID improves from 57.53 to 39.05.
These results indicate that the proposed framework primarily improves perceptual fidelity and noise realism, with only a limited sacrifice in pixel-level distortion.

\noindent\textbf{Qualitative Comparison.}
Fig.~\ref{fig:vis_lol_v1} shows qualitative comparisons on LOL-v1 based on full-image observations.
Across diverse scenes, the ``+'' variants generally produce cleaner outputs with reduced noise and improved global clarity, while better preserving scene structures and details.
Indoor scene: Our method alleviates noise and color cast, recovering more faithful colors and sharper edges without introducing noticeable artifacts.
Extremely low-light scene: Our method improves overall brightness and contrast while retaining fine structures (e.g., textural cues), yielding a more natural enhancement and avoiding over-smoothing or noisy amplification commonly observed in the baselines.

\subsubsection{Evaluation on Real-LOL-Blur}
\label{subsec:exp_real}
To evaluate real-world generalization, we assess models on the Real-LOL-Blur benchmark. Given its realistic low-light and blur degradations, evaluation is performed using no-reference metrics (NIQE, BRISQUE, and PI; lower is better).

\noindent \textbf{Pretraining on LOL-Blur.}
As reported in Tab.~\ref{tab:lol-blur-real-lol-blur}, our framework consistently improves baselines on Real-LOL-Blur, reducing NIQE, BRISQUE, and PI across all compared methods. This demonstrates effective transferability beyond the training distribution, leading to more natural and perceptually pleasing results under real degradations. Notably, stronger joint-enhancement backbones (e.g., Restormer, RetinexMamba, and CWNet) show particularly large gains.

\noindent \textbf{Pretraining on LOL-v1.}
Tab.~\ref{tab:lol-blur-real_v1} presents results for models pretrained on LOL-v1 and evaluated on Real-LOL-Blur. Despite the domain gap between LOL-v1 and real scenes, our framework still improves NIQE and PI for all baselines and significantly reduces BRISQUE for most. This demonstrates robust cross-dataset generalization. One exception is Retinexformer+, which shows a slight BRISQUE increase despite improving NIQE and PI. This suggests our framework primarily enhances global naturalness and perceptual fidelity even under challenging domain shifts. Overall, the consistent improvements across both pretraining sources underscore the practicality of our method in real-world scenarios.

\begin{table}[!t]
    \renewcommand{\arraystretch}{1.0}
    \setlength{\tabcolsep}{2pt}
    \centering
    \caption{Ablation study on modules.}
     \vspace{-0.2cm}
    \begin{tabular}{l|cccc}
        \toprule
        \textbf{Settings} & PSNR $\uparrow$ & SSIM $\uparrow$ & LPIPS $\downarrow$ & FID $\downarrow$ \\
        \midrule
        FourLLIE (Baseline)              & 20.06 & 0.8401 & 0.2409  & 88.17 \\\midrule
        w/o Noise Predictor & 20.83 & 0.8675 & 0.2277 & 84.95 \\
        \midrule
        w/o Gaussian-noise Expert & 20.79 & 0.8723 & 0.2173 & 83.91 \\
        w/o Poisson-noise Expert  & 21.12 & 0.8746 & 0.1857 & 79.74 \\
        w/o Gate                 & 20.89 & 0.8788 & 0.1962 & 80.80 \\
        w/o SFT layer            & 21.25 & 0.8801 & 0.1745 & 78.62 \\
        \midrule
        w/o PDS                 & 20.31 & 0.8493 & 0.2018 & 80.39 \\
        \midrule\rowcolor{gray!15}
        FourLLIE+ (Ours)              & \textbf{21.34} & \textbf{0.8815} & \textbf{0.1524}  & \textbf{76.87} \\
        \bottomrule
    \end{tabular}
     \vspace{-0.4cm}
    \label{tab:ablation_lolv1}
\end{table}

\begin{table}[!t]
    \renewcommand{\arraystretch}{1.0}
    \setlength{\tabcolsep}{2pt}
    \centering
    \caption{Ablation of loss terms.}
     \vspace{-0.2cm}
    \begin{tabular}{l|cccc}
        \toprule
        \textbf{Variant} & PSNR $\uparrow$ & SSIM $\uparrow$ & LPIPS $\downarrow$ & FID $\downarrow$ \\
        \midrule
        FourLLIE (Baseline)              & 20.06 & 0.8401 & 0.2409  & 88.17 \\\midrule
        w/o $\mathcal{L}_{\text{metric}}$ & 20.24 & 0.8459 & 0.2225 & 83.46 \\
        w/o $\mathcal{L}_{\text{consist}}$ & 21.05 & 0.8774 & 0.1857 & 79.74 \\
        w/o $\mathcal{L}^{\text{normal}}_{\text{np}}$ & 21.25 & 0.8749 & 0.1851 & 78.67 \\
        w/o $\mathcal{L}^{\text{low}}_{\text{np}}$ & 20.87 & 0.8698 & 0.1873 & 79.91 \\
        \midrule \rowcolor{gray!15}
         FourLLIE+ (Ours)              & \textbf{21.34} & \textbf{0.8815} & \textbf{0.1524}  & \textbf{76.87} \\
        \bottomrule
    \end{tabular}
     \vspace{-0.4cm}
    \label{tab:ablation_loss_terms}
\end{table}

\begin{table}[!t]
    \renewcommand{\arraystretch}{1.0}
    \setlength{\tabcolsep}{2pt}
    \centering
    \caption{Sensitivity to loss weights on LOL-v1.}
     \vspace{-0.2cm}
    \begin{tabular}{c|cccc}
        \toprule
        \textbf{Weight} & PSNR $\uparrow$ & SSIM $\uparrow$ & LPIPS $\downarrow$ & FID $\downarrow$ \\
        \midrule
         Baseline          & 20.06 & 0.8401 & 0.2409  & 88.17 \\\midrule
        $\lambda_{\text{con}}=0.3$ & 20.46 & 0.8752 & 0.2085 & 81.03 \\
        $\lambda_{\text{con}}=0.7$ & 21.37 & 0.8803 & 0.1840 & 77.56 \\
        \midrule
        $\lambda_{\text{met}}=0.001$ & 20.91 & 0.8560 & 0.1985 & 79.04 \\
        $\lambda_{\text{met}}=0.05$  & 21.32 & 0.8801 & 0.1842 & 78.58 \\
        $\lambda_{\text{met}}=0.1$   & 20.74 & 0.8785 & 0.1969 & 80.87 \\
        \midrule \rowcolor{gray!15}
        Default  ($\lambda_{\text{con}}=0.5$, $\lambda_{\text{met}}=0.01$)        & \textbf{21.34} & \textbf{0.8815} & \textbf{0.1524}  & \textbf{76.87} \\
        \bottomrule
    \end{tabular}
     \vspace{-0.4cm}
    \label{tab:ablation_loss_weights}
\end{table}

\subsubsection{Generalization to Real-World Deblurring and Denoising}
\label{subsubsec:exp_generalization}

To further evaluate the generalization of our framework on isolated real-world degradations, we extend our experiments to dynamic scene deblurring (GoPro~\cite{nah2017deep}) and extreme low-light denoising (SID~\cite{chen2019seeing}), using Retinexformer, Wave-Mamba, and CWNet as baselines.
As shown in Tab.~\ref{tab:gopro_sid}, integrating our framework yields consistent improvements across both datasets. On GoPro. Notably, CWNet+ gains $+0.37$\,dB in PSNR, while Retinexformer+ improves in SSIM despite a negligible PSNR drop ($-0.04$\,dB). On SID, characterized by extreme sensor noise, the performance leaps are even more pronounced across all models, with Retinexformer+ achieving a striking $+1.06$\,dB improvement in PSNR. These results compellingly validate that our framework generalizes well beyond joint low-light blur, serving as a robust plug-and-play solution for diverse real-world image enhancement tasks.

\subsection{Ablation Study and Analysis}
\label{subsec:ablation}

\begin{figure}[!t]
    \centering
    \includegraphics[width=\linewidth]{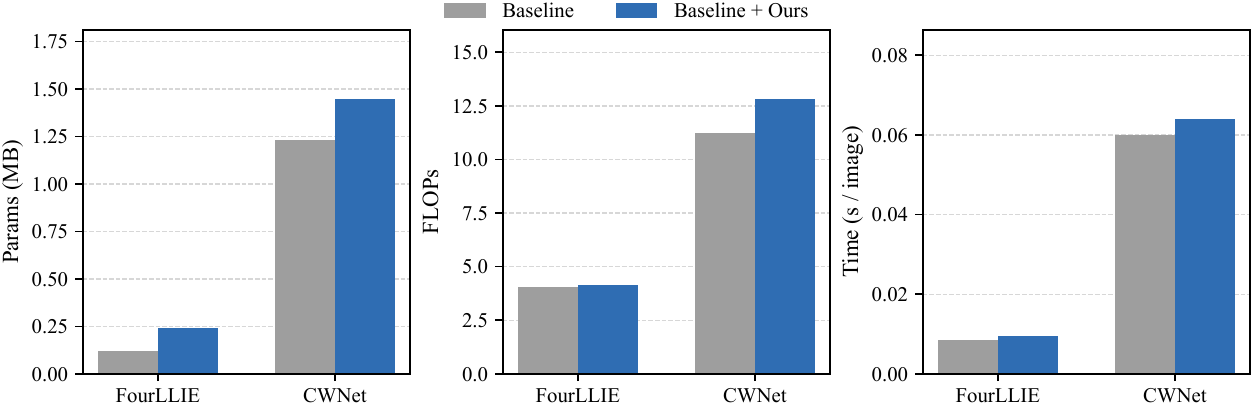}
     \vspace{-0.7cm}
    \caption{Efficiency comparison. We report Params, FLOPs, and average inference time per image for FourLLIE and CWNet.}
     \vspace{-0.3cm}
    \label{fig:eff_lolv1}
\end{figure}

\subsubsection{Contribution of Core Modules}
Tab.~\ref{tab:ablation_lolv1} summarizes the ablation results on LOL-v1.
Removing the noise predictor noticeably degrades performance, indicating that accurate noise estimation is important for stable enhancement.
For the proposed DA-MoE, disabling either the Gaussian-noise expert or the Poisson-noise expert leads to consistent performance drops. Further removing the gate also hurts performance, validating the necessity of expert specialization and adaptive routing.
The SFT layer provides a modest but consistent benefit.
Finally, removing PDS causes the most severe degradation, demonstrating that it is a key component for handling complex real degradations.

\subsubsection{Impact of Loss Functions}
\label{subsubsec:loss_ablation}

Tab.~\ref{tab:ablation_loss_terms} evaluates the impact of each loss term by removing it from the full objective.
All variants lead to performance drops, validating that each component contributes to enhancement quality.
In particular, removing the AMD loss causes the largest degradation, indicating that it plays a key role in guiding perceptual and structural fidelity.
Removing the dual-domain self-supervised loss and the two noise-predictor regularizations also consistently reduces PSNR/SSIM, demonstrating their complementary benefits for stable optimization and robustness.
Tab.~\ref{tab:ablation_loss_weights} further studies the sensitivity to loss weights.
For $\lambda_{\text{con}}$, an overly small weight degrades performance, while a larger weight (e.g., $0.7$) performs comparably to the default setting.
For $\lambda_{\text{met}}$, both too small and too large weights hurt performance, suggesting that a moderate value (default $\lambda_{\text{met}}{=}0.01$) provides a better balance between distortion and perceptual guidance.

\subsubsection{Computational Efficiency}
Most deep LLIE models operate at the scale of tens of MB parameters. To make the overhead of our framework more transparent under tight resource budgets, we choose two highly compact backbones, FourLLIE and CWNet, for efficiency analysis on LOL-v1.
As shown in Fig.~\ref{fig:eff_lolv1}, the parameter increase is small and remains well-controlled in absolute size: FourLLIE is extremely tiny (0.12\,MB), and our integration keeps its footprint at the 0.1\,MB level; CWNet increases from 1.23\,MB to 1.45\,MB (only +0.22\,MB, \(\sim\)17.9\%).
More importantly, the additional computation is marginal: the FLOPs grow only slightly, and the average inference time is nearly unchanged in practice (FourLLIE: 0.0085\,s \(\rightarrow\) 0.0096\,s; CWNet: 0.060\,s \(\rightarrow\) 0.064\,s per image), indicating that our gains are achieved with minimal efficiency overhead.

\section{Conclusion}
This paper tackles real-world LLIE by disentangling signal-dependent noise from high-frequency details in the sRGB domain, where ISP processing transforms physical noise into complex artifacts. The PDS stream inverts sRGB to a RAW proxy, injects physically plausible photon/read noise with explicit degradation vectors, and then re-projects to sRGB to synthesize high-fidelity training pairs. For enhancement, a noise predictor estimates degradation parameters to guide a DA-MoE for dynamic expert routing, while an AMD calibrates representation learning based on noise severity. For future work, we aim to explore additional real-world interference factors beyond Gaussian/Poisson noise, integrating them into a physically controllable degradation process for explicit adversarial optimization.

% \section*{Acknowledgments}
% This should be a simple paragraph before the References to thank those individuals and institutions who have supported your work on this article.


%\begin{thebibliography}{1}
\bibliographystyle{IEEEtran}
\bibliography{reference}

\vfill

\end{document}